# Image retrieval method based on CNN and dimension reduction

Zhihao Cao[1], Shaomin MU[1], Yongyu XU[2], Mengping Dong[1]
College of Information Science and Engineering, Shandong Agricultural University[1]
College of Plant Protection, Shandong Agricultural University[2]
Taian, shandong 271018, China
czh@sdau.edu.cn, msm@sdau.edu.cn, xuyy@sdau.edu.cn , dongmengping@126.com

*Abstract*— An image retrieval method based on convolution neural network and dimension reduction is proposed in this paper. Convolution neural network is used to extract high-level features of images, and to solve the problem that the extracted feature dimensions are too high and have strong correlation, multilinear principal component analysis is used to reduce the dimension of features. The features after dimension reduction are binary hash coded for fast image retrieval. Experiments show that the method proposed in this paper has better retrieval effect than the retrieval method based on principal component analysis on the e-commerce image datasets.

## I. INTRODUCTION

With the rapid development of e-commerce, the amount of image data on the electronic commerce platform has increased dramatically. Faced with the huge amount of image data, how to quickly and efficiently retrieve the commodity images that users are interested in has become an urgent problem to be solved [1].

Image retrieval first requires feature representation of images. The traditional image feature representation such as SIFT needs a priori knowledge to extract manually and the workload is heavy but the effect is unsatisfactory. Deep learning can automatically extract image feature [2], which has more advantages than traditional artificial features. At present, feature representation based on depth learning is widely used in many fields, such as text processing, speech recognition, image classification and target detection [3- 9].

Convolutional neural network (CNN) is a common depth learning model, which is widely used in the automatic feature extraction of images. It can form abstract high-level features by combining single features at the bottom of the image. The dimension of the features extracted by this method is very high, and some of the features have strong correlation. CNN is mainly composed of convolution layer, pooling layer, fully connected layer and classifier. The features extracted by convolution layer in CNN structure are called feature map, which is a kind of third-order tensor. The feature map retains the original structure information of the image. CNN will process the extracted feature map into a vector and input it into the fully connected layer, and then classify it using classifiers such as Softmax classifier.

In recent years, with the deep learning getting more and more attention, some scholars began to study the application of CNN in image retrieval [10]. Krizhevsky et al. [11] used feature vectors from fully connected layer for image retrieval and demonstrated excellent performance on ImageNet. Lin K et al. [12] presented a simple yet effective learning framework to create the hash-like binary codes for fast image retrieval, which proved that CNN can extract image features effectively. Babenko et al. [13] proposed to use principal component analysis(PCA) and discriminative dimensionality reduction to compress CNN features, and achieved good results. Ren et al. [14] used PCA to reduce the dimension of CNN features and used it in image retrieval, which proved that the retrieval effect was better than CNN features without dimension reduction. However, the image itself and the feature map extracted by CNN is a kind of third-order tensor, each layer of feature map has a close relationship. Since traditional PCA can only reduce the dimension of the feature vector [15], processing the feature map tensor data into feature vector data for dimension reduction will destroy the feature representation structure of original data.

To solve this problem, we use MPCA (Multilinear Principal Component Analysis) instead of PCA to reduce the dimensionality of image features. MPCA is an extension of PCA on high-dimensional data [16]. This algorithm can directly reduce the dimensionality of tensors, and achieve the purpose of dimension reduction through the combination of relevant features in tensors. Using MPCA algorithm to directly reduce the dimensionality of feature map tensor data can avoid the "dimension disaster" [17], and can effectively retain the structure information of the original image. Eventually the features are encoded into binary hash codes after dimension reduction and compared with the Hamming distance between the target image codes and the image codes in the database to achieve the retrieval of the e-commerce image.

## II. RELATED WORK

### A. Convolution Neural Network

CNN is a common deep learning model. Original images can be input into CNN without complex pre-processing, which can avoid complex low-level feature extraction and data reconstruction. Therefore, CNN has been widely used in the field of image pattern recognition [18]. CNN is a multi-layer supervised learning network consisting of convolution layer, pooling layer, fully connected layer and classifier. The network structure is optimized and the network parameters are adjusted by back-propagation algorithm. The CNN structure is depicted in Figure 1.

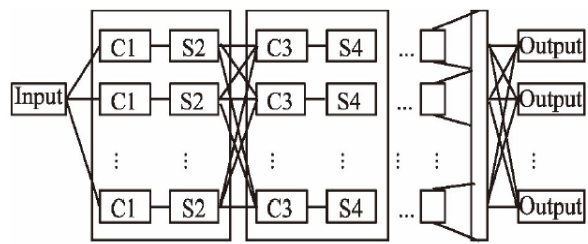

Fig. 1. Convolutional neural network structure diagram

## B. Multilinear Principal Component Analysis

The traditional PCA algorithm is to reduce the dimension of vectors [19]. MPCA is an extension of PCA on high-dimensional data. It directly reduces the dimension of high-order tensors [20]. The idea of the algorithm is to find multi-linear transformation $V_j (j = 1,2 \dots n)$ to map the original high-order tensor $X_i \in R^{m_1 \times m_2 \dots m_n}(i = 1,2 \dots N)$ to a low-order tensor $Y_i \in R^{l_1 \times l_2 \dots l_n}(i = 1,2 \dots N)$, N represents the total number of training samples and n represents the tensor order, $Y_i = X_i \times V_1^T \times V_2^T \dots \times V_n^T$, and to try keeping the scatter of the original tensor group unchanged.

The simplified MPCA algorithm is as follows:

1) Center the input samples:
$$\widetilde{X}_i = X_i - \overline{X} \quad (1)$$
where $\overline{X} = \frac{1}{N}\sum_{i=1}^{N} X_i$ is the sample mean.

2) Calculate k-mode total scatter matrix of the samples by
$$S^{(k)} = \sum_{i=1}^{N}(X_i^{(k)} - \overline{X}^{(k)})(X_i^{(k)} - \overline{X}^{(k)})^T \quad (2)$$
where $X_i^{(k)}$ is the $k$-mode unfolded matrix of $X_i$, and set $V_j$ to consist of the eigenvectors corresponding to the most significant $d_j$ eigenvalues.

3) The feature tensor after projection is obtained as
$$Y_i = X_i \times V_1^T \times V_2^T \dots \times V_n^T \quad (3)$$

## C. Locality-Sensitive Hashing

Locality-sensitive hashing is a common approximate nearest neighbor search algorithm. It maps high-dimensional floating-point data into binary codes, and then calculates Hamming distance between different binary codes, which reduces the calculation cost, simplifies the algorithm structure, and improves the retrieval speed [21]. The basic idea of LSH is that after the adjacent data points in the original space are mapped through the same mapping function, the probability that the two points are still adjacent in the new space is very high [22].

## III. PROPOSED IMAGE RETRIEVAL METHOD

In this paper, an image retrieval method which combines the advantages of features extraction method based on CNN, MPCA algorithm and LSH is proposed. The method model presented in this paper is shown in Figure 2.

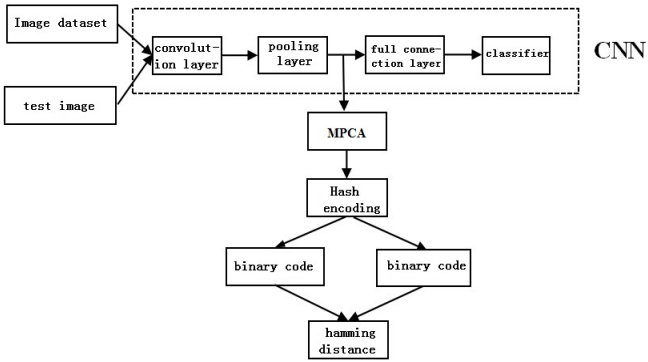

Fig. 2. Model diagram of Proposed Image Retrieval method

The feature extraction method based on deep learning is implemented by CNN. The CNN structure used in this paper is shown in Table 1. The network consists of five convolution layers, five pooling layers and three fully connected layers, and each of the convolution layers is connected to a activation layer using Relu. The input image size is 200×200×3, and the output size is 6×6×128 after the fifth convolution layer and the pooling layer, that is, the size of each feature map is 6 x 6, with 128 feature maps. After the first two fully connected layers, the output is a 1024-dimensional vector. After the last fully connected layer, a 5-dimensional vector is output and input to the Softmax classifier for classification.

TABLE I. CNN STRUCTURE PARAMETERS

| network layer | kernel size | kernel number | Output dimension |
|---|---|---|---|
| convolution layer 1 | 5×5 | 32 | 200×200×32 |
| pooling layer 1 | 2×2 | - | 100×100×32 |
| convolution layer2 | 5×5 | 32 | 100×100×32 |
| pooling layer 2 | 2×2 | - | 50×50×32 |
| convolution layer3 | 5×5 | 64 | 50×50×64 |
| pooling layer 3 | 2×2 | - | 25×25×64 |
| convolution layer4 | 3×3 | 128 | 25×25×128 |
| pooling layer 4 | 2×2 | - | 12×12×128 |
| convolution layer5 | 3×3 | 256 | 12×12×256 |
| pooling layer 5 | 2×2 | - | 6×6×256 |
| fully connected layer 1 | - | - | 1024 |
| fully connected layer 2 | - | - | 512 |
| fully connected layer 3 | - | - | 5 |

Traditional PCA algorithm inputs are vectors, so some researchers use the output vectors of fully connected layer as the inputs of PCA algorithm, that is, the corresponding 1024-dimensional vector in this CNN. MPCA algorithm can reduce the dimension of high-order tensors, so it is not needed to transform the high-order tensors into vectors. This algorithm will not destroy the spatial structure of the convoluted image, and can retain more effective information, so it can directly reduce the dimension of the feature map. In this paper, the output feature map of fifth pooling layer whose size is 6×6×256 are used as image features, and the dimensionality is reduced by MPCA algorithm.

The feature map of the image after centering can be expressed as $X_i \in R^{6 \times 6 \times 256}(i = 1,2 \dots N)$, the number $N$ of images used in this experiment is 80000. For the feature map $X$, it has three unfolded forms, namely 1-mode, 2-mode and 3-mode, and the 3-mode unfolded form is depicted in Figure 3.

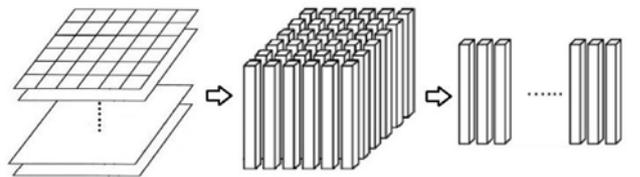

Fig. 3. 3-mode unfolded form of feature map

The total scatter matrixes in the three unfolded forms are:

$$S^{(1)} = \sum_{i=1}^{80000}(X_i^{(1)} - \bar{X}^{(1)})(X_i^{(1)} - \bar{X}^{(1)})^T \quad (4)$$

$$S^{(2)} = \sum_{i=1}^{80000}(X_i^{(2)} - \bar{X}^{(2)})(X_i^{(2)} - \bar{X}^{(2)})^T \quad (5)$$

$$S^{(3)} = \sum_{i=1}^{80000}(X_i^{(3)} - \bar{X}^{(3)})(X_i^{(3)} - \bar{X}^{(3)})^T \quad (6)$$

where $S^{(1)} \in R^{6\times 6}$, $S^{(2)} \in R^{6\times 6}$, $S^{(3)} \in R^{256\times 256}$, it can be seen that the dimension of the total scatter matrixes is low, so it is not necessary to use the singular value decomposition commonly used in PCA for high dimensional data, and the eigenvalue decomposition can be used directly.

After eigenvalue decomposing of $S^{(1)}, S^{(2)}$ and $S^{(3)}$, the corresponding eigenvalues and eigenvectors of the total scatter matrixes are obtained, and the eigenvalues are arranged in descending order. Then the eigenvectors corresponding to the first $d_1$, $d_2$ and $d_3$ eigenvalues are selected to form the projection matrixes $V_1 \in R^{6\times d_1}$, $V_2 \in R^{6\times d_2}$ and $V_3 \in R^{256\times d_3}$. According to the above steps of MPCA algorithm, we get the projection matrixes $V_1, V_2$ and $V_3$.

The image features after dimensionality reduction using MPCA can be expressed as $Y_i = X_i \times V_1^T \times V_2^T \times V_3^T$, where $V_1^T, V_2^T$ and $V_3^T$ are projection matrixes in three modes. The tensor after dimension reduction of MPCA algorithm is expanded to form a vector for binary hash coding. Then the tensor is spread into the matrix according to the 3-mode unfolded forms above, and then the matrix is spread into a vector in row.

Due to the large number of images, traditional linear search is not suitable for image retrieval in this paper. Hash algorithm can map multi-dimensional floating-point vector data into binary encoding. The representation of features is not only simplified, but also can be compared quickly by using Hamming distance in similarity measurement, which greatly speeds retrieval and improve real time performance

LSH is used for feature binary coding in this paper. According to the mapping result between floating-point data and multiple random mapping functions, the vector is encoded into $Hash(x) = [h_1(x), h_2(x), ... h_L(x)]$, where the value of $h_i(x)$ is 0 or 1. After hash coding, a picture can be represented by a shorter code.

All images in datasets is processed according to the above method, and their hash encodes is generated and saved to the database. When a query image is input, a corresponding hash code $Hash(x)$ is generated, and the Hamming distance of the hash codes between it and all images in the database is calculated. The shorter the distance, the higher the similarity between the two pictures is.

IV. EXPERIMENT

*A. Data and Environment*

The images in the experiment were obtained by crawler technology from Taobao and Tmall based on Python and divided into 10 categories: dress, half skirt, suits, furs coat, short sleeves, long sleeves, shorts, jeans, high heels and leather shoes. The size of the image is 200×200, 1000 images for each class, and part of the images is shown in Figure 4. In order to improve the training effect of CNN, the sample images were enhanced by translating, rotating, illumination changing and so on. Each image was made into several images, and the total number of images is 80000.

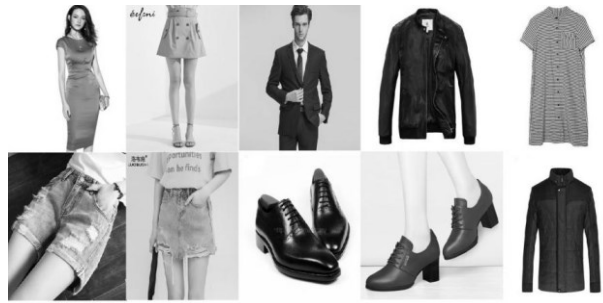

Fig. 4. Part of e-commerce images in experiment

The experiment is carried out on the high-performance computing platform of Shandong Agricultural University. CNN is programmed with TensorFlow 1.9 framework, MPCA algorithm is implemented with MATLAB open source codes, Hash coding is implemented with Lshash module in Python.

*B. Results and Analysis*

First of all, 10 classes of image data and corresponding labels are input into CNN for training. Stochastic gradient descent (SGD) is used to adjust the network parameters. The CNN learning rate is set to 0.0001 and the batch is set to 64. When the network iterates 30000 times, the network converges and the final classification accuracy is 91.2%, and the process of training is depicted in Figure 5.

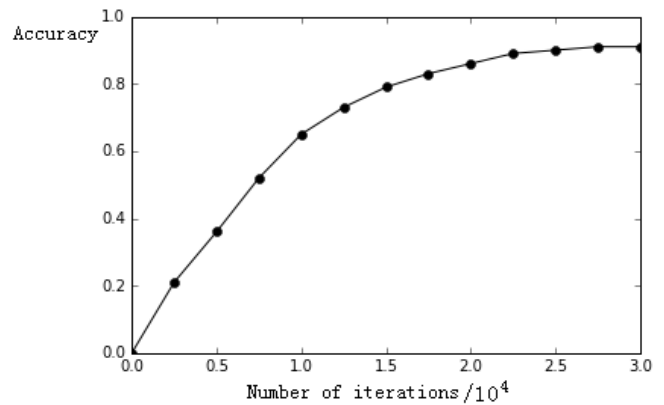

Fig. 5. The process of CNN training

Dimension reduction experiment is carried out based on MPCA with reference to Ref. [19]. In order to find the optimal feature dimension after dimension reduction, a number of different groups of $d_1, d_2$ and $d_3$ are selected for comparison. Since the dimension of each layer is 6×6, and the dimension in each direction is 6, the dimensionality after dimension reduction should not be too large or too small. When dimensionality after dimension reduction is too large, dimension reduction is meaningless. When dimensionality after dimension reduction is too small, many important features will be abandoned in the process. Therefore we chooses to reduce the dimensionality to 2, 3 and 4. In order to maintain the overall structure of the original feature, the same compression rate (CR) is chosen in three directions, where the compression rate is equal to (the number of features after compression) / (the number of original features). The final three groups of selected numbers of feature dimensionality in the experiment are shown in Table 2.

TABLE II. SELECTED NUMBERS OF FEATURE DIMENSIONALITY

| No. | $d_1$ | $d_2$ | $d_3$ | compression rate |
|-----|-------|-------|-------|------------------|
| 1 | 2 | 2 | 85 | 33.3% |
| 2 | 3 | 3 | 128 | 50.0% |
| 3 | 4 | 4 | 170 | 66.7% |

In the three groups of experiments, the original feature maps $X \in R^{6\times 6\times 256}$ are respectively dimension reduced to $X_{(1)} \in R^{2\times 2\times 85}, X_{(2)} \in R^{3\times 3\times 128}$ and $X_{(3)} \in R^{4\times 4\times 170}$. The experimental results are evaluated with reference to the related concepts of principal component analysis algorithm. The principal components are calculated through eigenvalue decomposition of total scatter matrixes in the three modes, and they are in descending order according to eigenvalue. The higher the contribution rate of principal components is, the higher the ranking is. The cumulative contribution rates (CCR) of the first $d_1, d_2$ and $d_3$ principal components in the three groups of experiments were statistically analyzed, and they are shown in Table 3.

TABLE III. CUMULATIVE CONTRIBUTION RATES

| no | CR | $S^{(1)}$ | | $S^{(2)}$ | | $S^{(3)}$ | |
|----|----|-----------|------|-----------|------|-----------|------|
| | | $d_1$ | CCR | $d_2$ | CCR | $d_3$ | CCR |
| 1 | 33.3% | 2 | 85.2% | 2 | 85.4% | 85 | 91.4% |
| 2 | 50.0% | 3 | 90.6% | 3 | 90.9% | 128 | 95.1% |
| 3 | 66.7% | 4 | 92.5% | 4 | 92.7% | 170 | 97.5% |

As can be seen from Table 3, the cumulative contribution rates of the first $d_1$ and $d_2$ principal components in $S^{(1)}$ and $S^{(2)}$ are very similar, indicating that the features in both directions of the same layer feature map have the same status. Because the images in the dataset are enhanced by rotation, the original image and the rotated image are both input to CNN for training, and the length and width of the image are exactly the same, the two directions of images have equal status in CNN training, that is, length and width. The rotation of the image is shown in Figure 6.

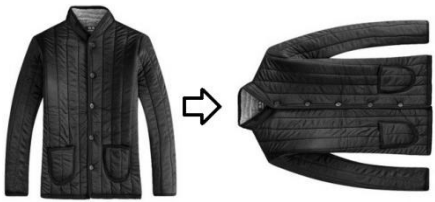

Fig. 6. The rotation of the image

When the original image data is compressed by 50%, the cumulative contribution rates of principal components in the three directions are as high as 90.6%, 95.1% and 90.9%, and still as high as 85.2%, 85.4% and 91.4% when the original data is compressed to 33.3%. It shows that the features extracted by CNN have strong correlation, and some of them have low contribution to image recognition.

The cumulative contribution rates are calculated from three directions above. Then the cumulative contribution rates of the three groups of principal components are weighted according to the proportion of the number of features, which can measure the cumulative contribution rate on the whole scale. Assuming that the cumulative contribution rates respectively of the first $d_1, d_2$ and $d_3$ principal components are $CCR_1$, $CCR_2$ and $CCR_3$, the weighted cumulative contribution rates of principal components can be calculated by

$$CCR_w = CCR_1 \frac{6}{6+6+256} + CCR_1 \frac{6}{6+6+256} + CCR_1 \frac{256}{6+6+256} \quad (9)$$

The weighted cumulative contribution rates of principal components of the three groups of experiments are shown in Table 4.

TABLE IV. WEIGHTED CUMULATIVE CONTRIBUTION RATES

| No. | $CCR_w$ |
|-----|---------|
| 1 | 91.1% |
| 2 | 94.9% |
| 3 | 97.2% |

In order to compare with effect of the PCA algorithm, we use PCA to reduce the dimensionality of the second group of fully connected layer output vector $f \in R^{1024}$, and select the same cumulative contribution rate as above. That is to say, in the PCA algorithm, the eigenvectors corresponding to the $i$ largest eigenvalues $\lambda_1, \lambda_2 \dots \lambda_i$ of the covariance matrix are selected as the principal components, and the cumulative contribution rates of this principal components are 91.1%, 94.9% and 97.2% respectively. The $i$ is calculated to be 58, 72 and 128 respectively, and the $f \in R^{1024}$ is dimension reduced to $f_1 \in R^{1024}$, $f_2 \in R^{1024}$ and $f_3 \in R^{1024}$. The parameters of dimension reduction are shown in Table 5.

TABLE V. PARAMETERS OF DIMENSION REDUCTION

| No. | original dimension | present dimension | CCR |
|-----|-------------------|-------------------|-----|
| 1 | 1024 | 231 | 91.1% |
| 2 | 1024 | 288 | 94.9% |
| 3 | 1024 | 512 | 97.2% |

The feature map tensor after dimension reduction based on MPCA and the feature vector output from fully connected layer after dimension reduction based on PCA are hashing coded respectively. Two hash codes with different lengths are used in the experiment, and the length is 64 and 128 commonly used in hash encoding. Mean average precision (MAP) is used to evaluate the retrieval results. For three groups of different cumulative contribution rates and two hash codes of different lengths, two dimension reduction methods are compared, and the results are depicted in Figure 7.

As can be seen from Figure 7, as for both the methods based on MPCA and PCA, the retrieval effect using 128-bit hash coding have a significant improvement compared to the 64-bit hash coding. Because of the large number of images, more features can distinguish the images more accurately. However, the codes is longer and the retrieval speed will be reduced.

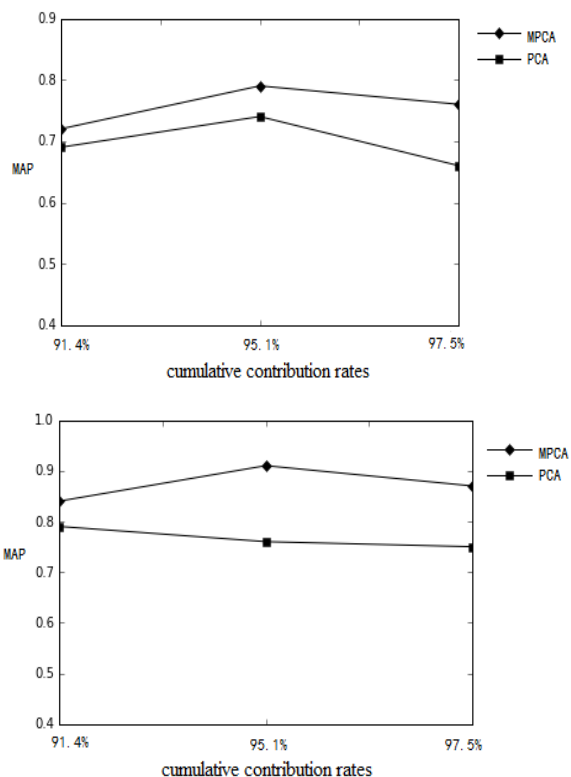

Fig. 7. experimental results

Under the same cumulative contribution rate, MAP of MPCA-based retrieval method has a significant improvement compared with PCA-based retrieval method. When the 128-bit hash encoding is used and the feature map is dimension reduced to $X \in R^{3 \times 3 \times 128}$, the MAP reaches the maximum value of 90.1%. Experiments show that MPCA is more effective in dimension reduction of image feature map than PCA. MPCA can retain more image structure information and improve the retrieval accuracy compared with PCA. At the same time, it is proved that there is a strong correlation between the features extracted by CNN. Using the new features after dimension reduction to represent the image can not only effectively express the information contained in the image, but also effectively improve the speed of retrieval in the era of big data.

## V. Conclusion

In this paper, an image retrieval method based on deep learning and dimension reduction is proposed. High-level features of images are extracted by CNN, and then the dimensionality of extracted features is reduced by MPCA. After dimension reduction, hash coding is used for image retrieval. Experiments show that the features extracted by CNN have strong correlation and are not suitable for direct image coding. The dimension reduction of feature map by MPCA is better than that by PCA. However, CNN is insensitive to the spatial relationship between objects in images. In the future, we will use CapNet to extract image features, and do some related theoretical research.

## Acknowledgment

This work is supported by First Class Discipline Funding of Shandong Agricultural University and The Modern Tea Industry Technology System of Shandong Province，China (SDAIT-19-04).